\documentclass[runningheads]{llncs}
\usepackage[T1]{fontenc}
\usepackage[dvipsnames]{xcolor}
\usepackage{graphicx}
\usepackage{tikz}
\usepackage{subcaption}
\usepackage{float}
\usepackage{amsmath}
\usepackage{amssymb}
\usepackage{tabularx}
\usepackage{bm}
\usepackage{tabularray}
\usepackage{array} 
\usepackage{algorithm}
\usepackage{algpseudocode}
\usepackage{caption}
\usepackage[colorlinks=true, urlcolor=blue, linkcolor=red]{hyperref}
\begin{document}
%
\title{HUP-3D: A 3D multi-view synthetic dataset for assisted-egocentric hand-ultrasound pose estimation}
%
\titlerunning{HUP-3D: A synthetic 3D hand-ultrasound dataset}
%
%
%

\author{Manuel Birlo \and
Razvan Caramalau \and
Philip J. ``Eddie'' Edwards \and
Brian Dromey \and
Matthew J. Clarkson \and
Danail Stoyanov}

\authorrunning{M. Birlo et al.}

\institute{Wellcome/EPSRC Centre for Interventional and Surgical Sciences (WEISS), 
University College London, 
Charles Bell House, 43–45 Foley Street, \linebreak London W1W 7TY, UK\\
\email{manuel.birlo.18@ucl.ac.uk}
}

\maketitle              
\begin{abstract}

We present HUP-3D, a 3D multi-view multi-modal synthetic dataset for hand-ultrasound (US) probe pose estimation in the context of obstetric ultrasound. 
Egocentric markerless 3D joint pose estimation has potential applications in mixed reality based medical education. The ability to understand hand and probe movements programmatically opens the door to tailored guidance and mentoring applications.   
Our dataset consists of over 31k sets of RGB, depth and segmentation mask frames, including pose related ground truth data, with a strong emphasis on image diversity and complexity. Adopting a camera viewpoint-based sphere concept allows us to capture a variety of views and generate multiple hand grasp poses using a pre-trained network. Additionally, our approach includes a software-based image rendering concept, enhancing diversity with various hand and arm textures, lighting conditions, and background images. 
Furthermore, we validated our proposed dataset with state-of-the-art learning models and we obtained the lowest hand-object keypoint errors.
The dataset and other details are provided with the supplementary material. The source code of our grasp generation and rendering pipeline will be made publicly available.

\end{abstract}

\section{Introduction}

\label{sec:intro}
The ability to infer hand and tool pose information from video data in clinical setups opens the door to several potentially useful applications aimed at assisting clinicians through context-specific evaluation of their movement behavior, and respective novel methods focusing on marker-free video-based clinical skill assessment were already proposed such as surgical hand and tool pose estimation~\cite{goodman2021real}, surgical tool movement analysis~\cite{Jin2018ToolDetection} and skill assessment in robotic surgery~\cite{lajko2021endoscopic}. Furthermore, solutions for skill assessment that rely on physical motion sensors have been developed in fields such as hand motion analysis for endovascular procedures~\cite{goodman2021real} and guidance for using ultrasound (US) probes in obstetric US~\cite{Jin2018ToolDetection}.
With the emerging trend of mixed reality head mounted displays, dominated by the Microsoft HoloLens 2~\footnote{\url{https://www.microsoft.com/en-us/hololens}}, egocentric pose estimation methods arose, such as probe tracking for US-guided procedures~\cite{nguyen2022holous}.

In the context of obstetric US, we explore 3D joint hand-tool pose estimation with an eye toward applications in mixed reality-based medical education, where analysis of hand and probe movements could facilitate holographic assistive probe guidance. Such innovations aim to help standardize clinical training protocols, addressing the current lack of universally accepted competence measures for effective US probe guidance, despite the existence of standardized target scanning US planes~\cite{CAI2020101762}. By distinguishing between novice and expert clinician movements, machine learning-based pose estimation emerges as a powerful tool for developing standardized training approaches. This technique not only aligns with but also supports established clinical practices for fetal weight estimation through biometry, offering a pathway to more uniform and effective clinical training methodologies~\cite{Dromey2020DimensionlessUltrasound}. 

Image datasets required for machine learning-based model training and subsequent pose estimation can be categorized into real and synthetic images.
Real dataset generating methods often employ marker-free methods to capture hand grasp information directly from image data~\cite{hasson19_obman,Hein2021,wang2023pov,GRAB:2020,Hampali2020HOnnotate}. While real images offer the advantage of authentic contexts~\cite{Hampali2020HOnnotate,kwon2021h2o}, they pose challenges for generating accurate ground truth due to the labor-intensive nature of manual annotations and potential biases from sensor use~\cite{Hein2021}. Marker-free approaches, by avoiding markers on the tool and/or hand, mitigate the risk of pose prediction bias, making them a popular choice for reducing inaccuracies associated with additional visible sensors or markers~\cite{wang2023pov,Hein2021}.

Synthetic images, however, offer built-in ground truth from 3D models, simulating realistic grasping scenarios with the benefits of easy scalability and generalizability to real images~\cite{hasson19_obman}. Furthermore, synthetic ground truth proves useful for addressing mutual occlusions resulting from hand-tool interactions.

When creating training images for pose estimation, it's crucial to account for the dataset's generalizability, particularly the expected camera location and range of viewpoints. Applications vary, with some utilizing non-egocentric views for capturing hand grasps~\cite{hasson19_obman,Hampali2020HOnnotate}, while mixed and augmented reality setups, especially those using head-mounted devices like the Microsoft HoloLens 2\footnote{\url{https://www.microsoft.com/en-us/hololens/}}, require considering egocentric perspectives for pose estimation from device-recorded camera data~\cite{Hein2021,wang2023pov}.
When capturing clinical instruments like US probes in synthetic images, the realism of hand grasps is constrained by specific hand-tool contact areas and orientations. This requirement, previously noted in the context of orthopedic surgical tools such as drills~\cite{Hein2021} and other instruments like scalpels and diskplacers~\cite{wang2023pov}, poses a challenge. Traditional grasp generation techniques, like those enabled by robotic grasping software~\cite{miller2004graspit} and used in similar studies~\cite{hasson19_obman,Hein2021}, encounter difficulties due to the specific dimensions and clinically relevant grasp positions of the US probe. Consequently, we employed more flexible solutions such as a generative model for machine learning-based grasp generation~\cite{GRAB:2020}, which has been validated in clinical environments~\cite{wang2023pov}.

To increase image diversity for egocentric applications in a scalable way, we broadened the approach to multi-view by allowing camera movement around a sphere's surface, centered on the hand. This method supports both egocentric distances and a mix of egocentric and non-egocentric viewpoints.

Our contributions can be summarized as follows:
\begin{itemize}
    \item A scalable synthetic multi-modal (RGB-D, segmentation maps) image generation pipeline capable of producing a wide variety of realistic hand-ultrasound probe grasp frames, independent of prior external data recording requirements
    \item A novel sphere-based camera viewpoint concept that enhances frame generalizability by combining egocentric head-hand distances with non-egocentric camera viewpoints. 
    \item HUP-3D: A diverse multimodal synthetic dataset tailored for joint 3D hand and tool pose estimation, featuring the Voluson™ C1-5-D\footnote{\url{https://services.gehealthcare.com/gehcstorefront/p/5499513}} ultrasound probe commonly used in obstetrics, including a variety of hand poses, textures, backgrounds, lighting, and camera angles
    \item Lowest hand and object 3D pose estimation errors for a synthetic dataset with a trained state-of-the-art model, HOPE-net \cite{hope}
\end{itemize}



\section{Method}
\begin{figure}[!htb]
\includegraphics[width=\textwidth]{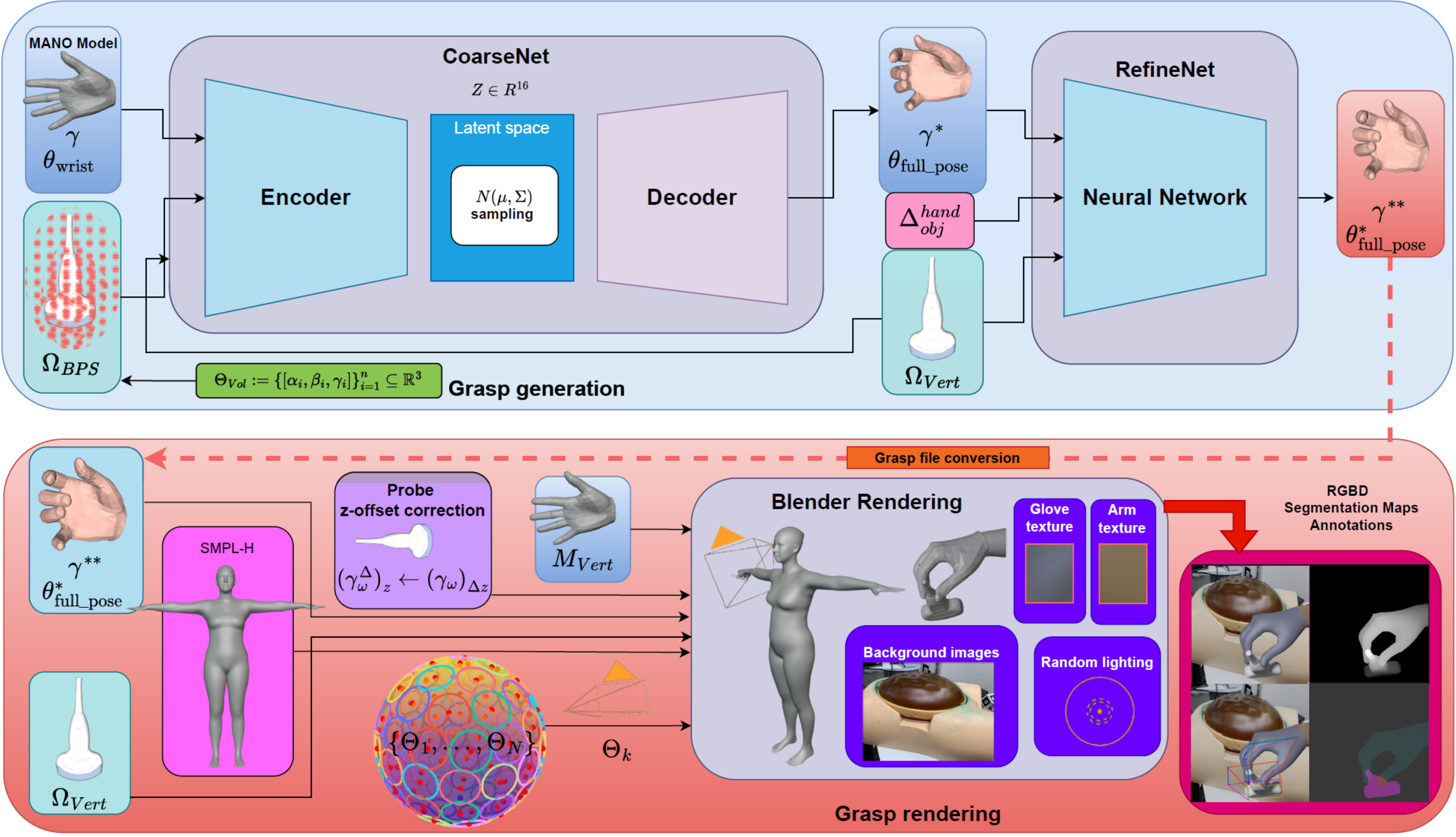}
\vspace*{-6mm}
\caption{
Grasp Generation (blue) and Rendering Pipeline (red): The process begins with a MANO hand model initialization and a BPS-encoded Voluson model point cloud. CoarseNet generates initial hand poses, further refined by RefineNet for precise hand-probe alignment. In the rendering phase, the optimized hand pose, model vertices, and a SMPL-H model are processed in Blender. Using a multi-viewpoint camera via a spherical layout and centered on the hand and arm, several textures and backgrounds are applied for diverse RGB-D, segmentation maps, and annotations.
}
\label{fig:pipeline_overview}
\end{figure}




Focusing on potential medical education applications in the context of obstetric US, but with a high degree of flexibility towards other use cases, as described in section~\ref{sec:intro}, our synthetic image generation pipeline is split into two sections: grasp generation and grasp rendering. These are described in detail in the following subsections. A graphical overview of our pipeline is shown in Fig. \ref{fig:pipeline_overview}.

\subsection{Grasp generation}
\label{sec:grasp_generation}
To streamline ground truth annotation, we adopted a strategy focused on generating synthetic grasp images, avoiding the complexities associated with annotating real images. This approach allowed us to maintain a clear and manageable rendering workflow. 
The underlying idea in pursuing a purely synthetic image generation approach is to explore the possibility of generating a sufficiently large variety of training images that allow acceptable generalizability to real images in joint 3D hand and tool pose prediction.
Our initial feasibility study incorporated the use of the robotic grasping tool~\cite{miller2004graspit} already mentioned in section~\ref{sec:intro}, which turned out to be error-prone and did not produce a sufficiently large variety of plausible hand grasps due to restrictions in terms of hand dexterity. Therefore, we adapted the generative model proposed in~\cite{GRAB:2020} for joint 3D grasp generation to a more clinical scenario.
Our grasp generation process employs two sequential networks based on the MANO hand model~\cite{MANO:SIGGRAPHASIA:2017}: an encoder-decoder network that generates initial coarse hand poses and a subsequent neural network dedicated to fine-tuning these poses, specifically enhancing accuracy in hand-tool interaction areas. 
The encoder which samples from a normal distributed 16 dimensional latent space, requires encoded point cloud representations~\cite{prokudin2019efficient} of the probe model ($\Omega_{BPS}$), together with the MANO right hand model's initial translation $\gamma \in \mathbb{R}^{3}$ and hand wrist orientation $\theta_{wrist} \in \mathbb{R}^{3}$. Defined Euler angles $\Theta_{Vol}$ for probe meshes $\Omega_{BPS}$ were used for precise grasp pose control.
Originally, the model described in~\cite{GRAB:2020} was trained with ordinary objects (like mugs, cameras etc.). However, we extended its capability to the Voluson US probe
The decoder outputs an initial hand pose $[\gamma^{}, \theta_{full\_pose}]$, which is subsequently refined through a neural network utilizing the vertices of the probe model $\Omega_{Vert}$ and the vertex distances $\Delta^{hand}_{obj}$ between hand and probe. This refined pose, expressed as $\Psi := [\gamma^{**}, \theta^{}_{full\_pose}]$, forms the foundation for our grasp rendering approach detailed in Sec. \ref{section:grasp_rendering}.

\begin{figure}[!t]
\centering
\begin{subfigure}[b]{0.2\textwidth}
   \centering
   \begin{tikzpicture}
       \node[inner sep=0pt] (image1) {\includegraphics[width=\textwidth]{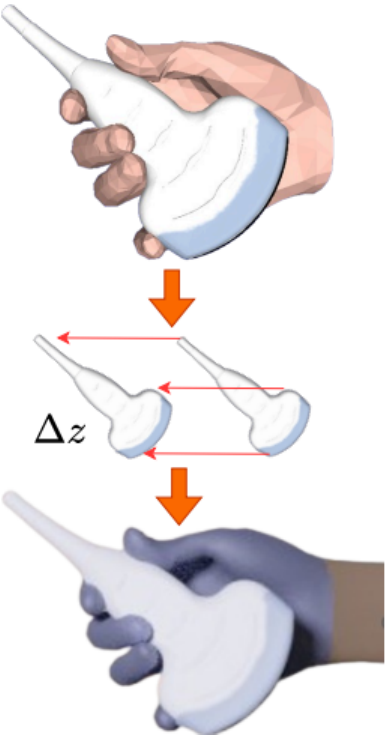}};
       \draw[orange,thick] (image1.south west) rectangle (image1.north east);
       \node[anchor=south west, fill=white, opacity=0.5, text opacity=1, inner sep=2pt] at (image1.south west) {(a)};
   \end{tikzpicture}
   \label{fig:grasp_comparison}
\end{subfigure}
\begin{subfigure}[b]{0.79\textwidth}
        \centering
        \begin{tikzpicture}
            \node[inner sep=0pt] (image2) {\includegraphics[width=\textwidth]{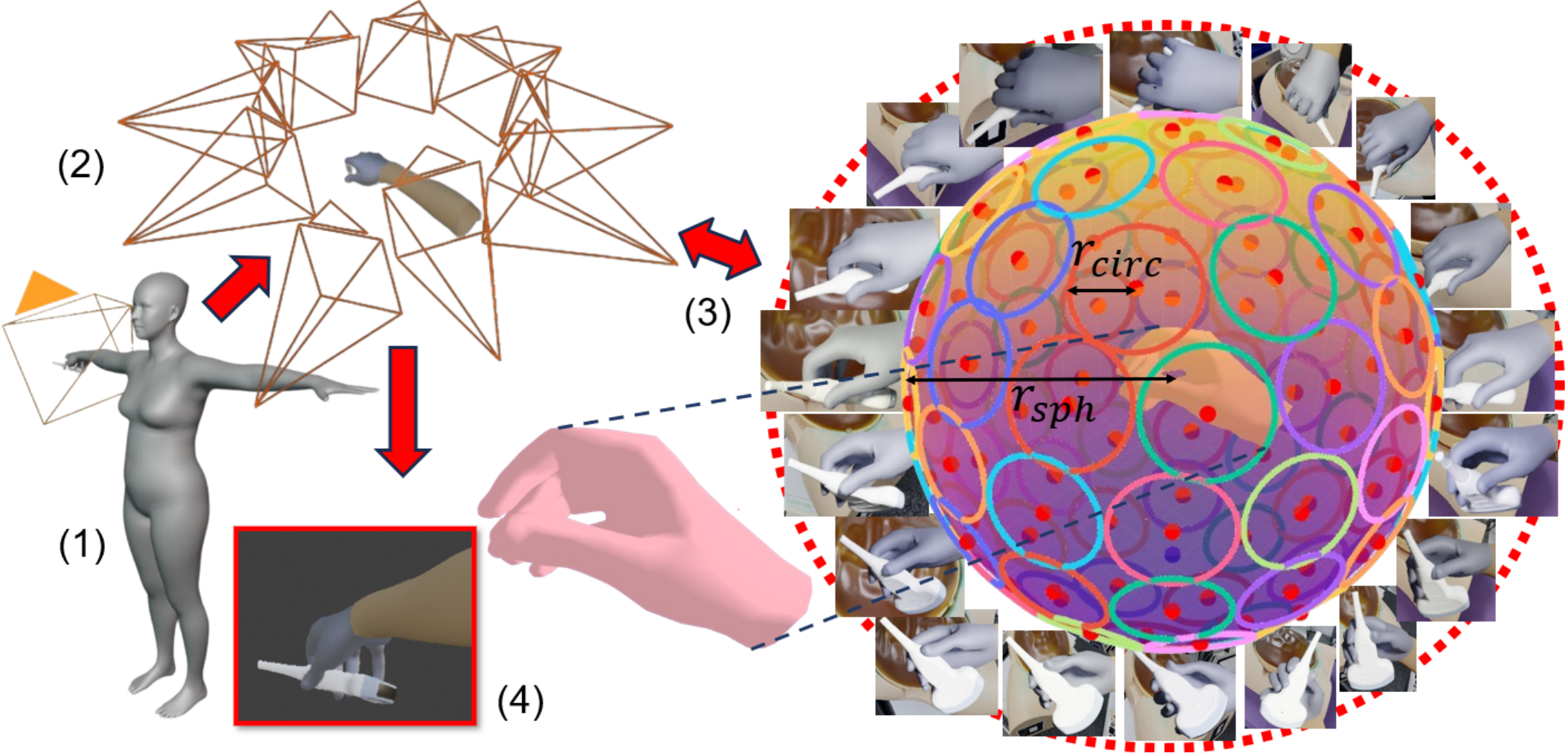}};
            \draw[orange,thick] (image2.south west) rectangle (image2.north east);
            \node[anchor=south west, fill=white, opacity=0.5, text opacity=1, inner sep=2pt] at (image2.south west) {(b)};
        \end{tikzpicture}
        \label{fig:camera_sphere_details_diagram}
\end{subfigure}
\caption{(a) Schematic grasp conversion from generative model to rendering software, including probe offset ($\Delta z$) correction. (b) Grasp rendering overview: (1) SMPL-H body model grasping the probe, showing egocentric and non-egocentric views. (2) Right arm and sphere-based camera orientations with remaining SMPL-H body parts hidden. (3) Camera angle sphere concept with views at various latitudes, centered on hand mesh; defines sphere ($r_{sphr}$) and circle ($r_{circ}$) radii. (4) Rendered hand-probe scene example from a sphere camera position.}
\label{fig:grasp_comparison_and_camera_sphere_details_diagram}
\end{figure}
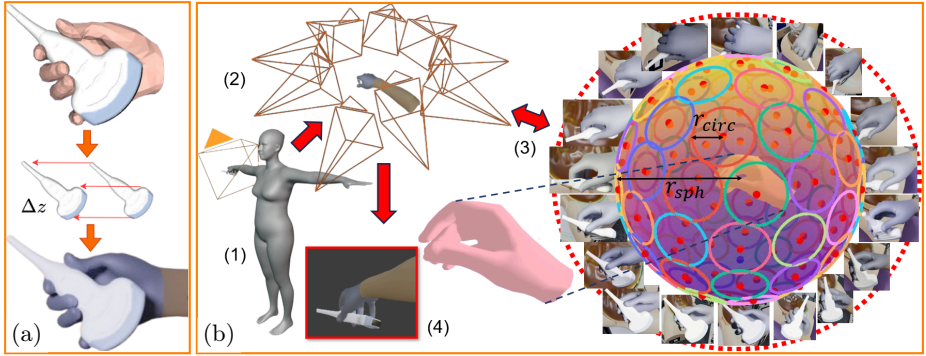

\footnotetext{\url{https://www.meshlab.net/}}

\subsection{Grasp rendering}
\label{section:grasp_rendering}
Building on the concept of utilizing the Blender~\cite{Blender2018} open-source 3D graphics software for grasp rendering, as demonstrated in~\cite{hasson19_obman,Hein2021}, we tailored our rendering pipeline to accommodate the grasp poses $\Psi$ produced by the generative model outlined in Sec. \ref{sec:grasp_generation}. Additionally, this rendering approach incorporates a SMPL-H body model~\cite{Romero2017EmbodiedHands}, a MANO right hand model $M_{Vert}$, and the probe model's vertex data $\Omega_{Vert}$.
Details of the grasp rendering part can be found in the lower part of Fig.~\ref{fig:pipeline_overview}.
Our methodology positions the probe at the origin of the rendering software's internal world coordinate system, yet the modified hand grasp pose $\Psi$ necessitates an offset for the probe model's translation $\gamma_{\omega}$, specifically a displacement of $\Delta z$ along the positive z-axis, which was calculated programmatically through polygon offset analysis.
Subsequently, we adjusted for the probe's translation offset via $\gamma^{\Delta}_{\omega} = \gamma_{\omega} + (0, 0, -\Delta z)$. Fig.~\ref{fig:grasp_comparison_and_camera_sphere_details_diagram}a schematically illustrates the probe's z-offset correction.
Fig.~\ref{fig:grasp_comparison_and_camera_sphere_details_diagram}a schematically illustrates the probe's z-offset correction.

To enhance the diversity of generated grasp images across different camera perspectives, we transitioned from the purely egocentric viewpoint strategy to a sphere-based methodology outlined in Sec. \ref{sec:sphere_concept}, illustrated in Figs.~\ref{fig:pipeline_overview} and \ref{fig:grasp_comparison_and_camera_sphere_details_diagram}b.
This helps a pose estimation model's capability to accurately predict grasp poses amid challenges like mutual occlusions between hands and tools.

\begin{figure}[t]
\centering
\includegraphics[width=.9\linewidth]{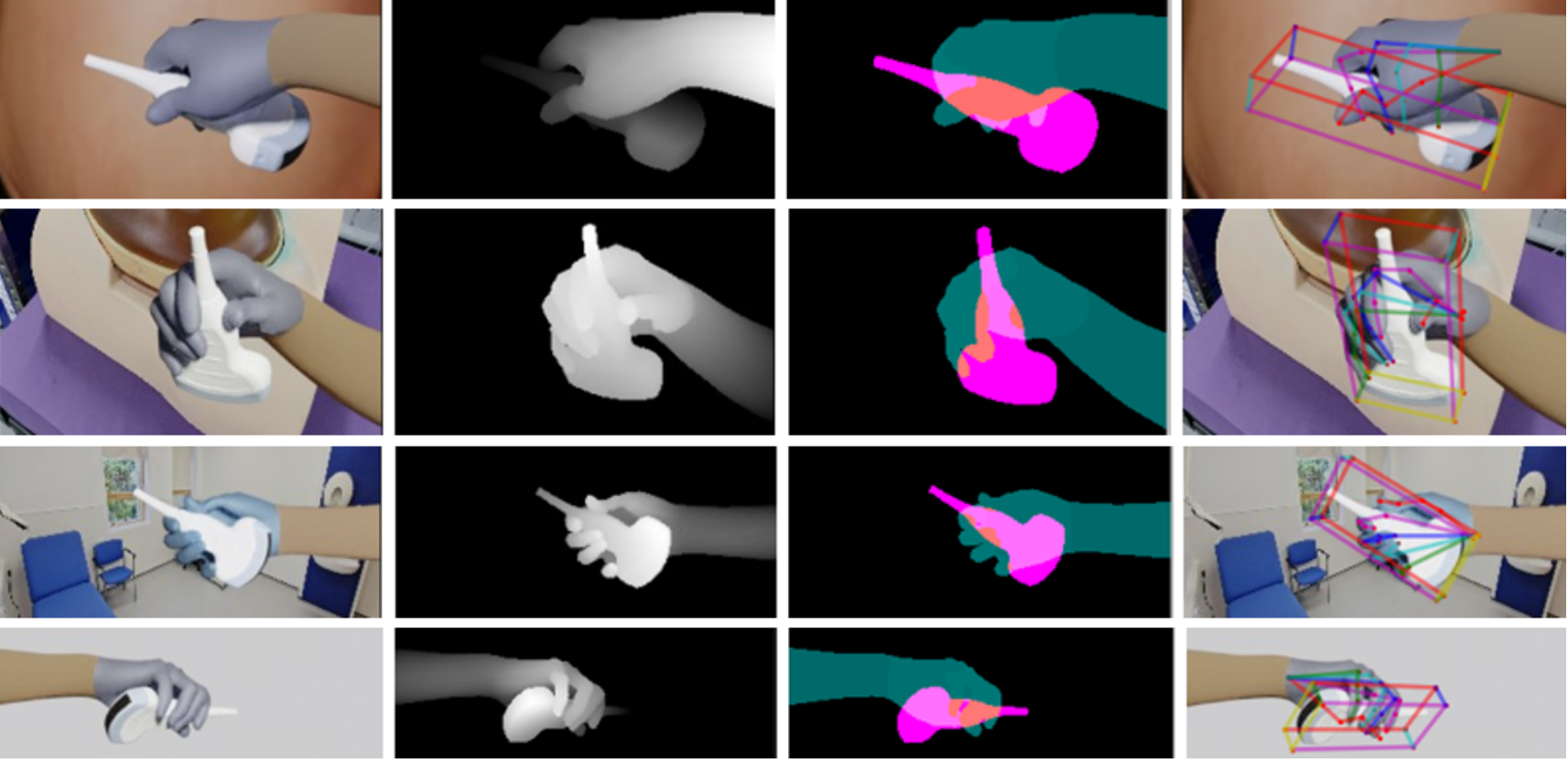}
\caption{Sample frames from the HUP-3D dataset, grouped columnwise, from left to right: RGB, depth, segmentation map, and ground truth annotations.}
\label{fig:rendered_images}
\end{figure}

For each grasp produced by our module, we generate a synthetic image for each camera view angle $\Theta_k \in \{\Theta_1, \ldots, \Theta_N\}$, covering $N$ positions around the sphere. Our render scene setup utilizes two shades of clinical gloves, varied scene lighting, and eight backgrounds (a lab with a SPACE-FAN ultrasound fetus model\footnote{\url{https://www.kyotokagaku.com/en/products\_data/us-7\_en/}}, consultation rooms, a white background, and real abdomens of pregnant women). The rendering model outputs a comprehensive set of images for each grasp, including RGB-D and segmentation maps, as well as ground truth annotations. Rendered sample frames from the HUP-3D dataset are shown in Fig.~\ref{fig:rendered_images}. 

\subsubsection{Camera view angle sphere concept}
\label{sec:sphere_concept}
Our methodology diverges from traditional egocentric viewpoints by implementing a sphere-based camera view setup to capture both egocentric and non-egocentric images, enhancing dataset diversity. This method, inspired by~\cite{Akin2012Enhanced}, involves distributing camera positions around a sphere, creating a varied perspective landscape around the right hand.
To systematize camera placement, the sphere is divided into horizontal segments, determined by latitude angles, to evenly distribute viewpoints. Specifically, the number of latitude segments \(N_{\text{latitude\_floors}}\) and circles per segment \(N_{\text{circles}}^{(i)}\) are calculated to ensure comprehensive coverage:
\begin{equation}
N_{\text{latitude\_floors}} = \left\lfloor \frac{\pi}{2 \arcsin\left(\frac{r_{circ}}{r_{sph}}\right)} \right\rfloor, \quad N_{\text{circles}}^{(i)} = \left\lfloor \frac{2 \pi r_{sph} \sin(\theta_i)}{2r_{circ}} \right\rfloor
\end{equation}
For each segment $i$ and each circle $j$ within, camera locations are defined by their spherical coordinates $(\theta_i, \phi_j^{(i)})$, ensuring a uniform spread of angles:
\begin{equation}
(\theta_i, \phi_j^{(i)}) \quad \text{with} \quad \phi_j^{(i)} = j \cdot \frac{2\pi}{N_{\text{circles}}^{(i)}}, \quad j = 0, 1, \dots, N_{\text{circles}}^{(i)} - 1
\end{equation}
This structured approach facilitates the generation of camera angles $\Theta_k$, utilized in our subsequent rendering process. Figs.~\ref{fig:pipeline_overview} and \ref{fig:grasp_comparison_and_camera_sphere_details_diagram}b visually demonstrate this concept, showcasing the strategic camera placement and the diverse grasp views it enables.

\subsubsection{Dataset comparison}
\label{sec:dataset_comparison}
In Table \ref{tab:hand_grasp_datasets}, we enlist the top clinical and non-clinical datasets, together with their properties. HUP-3D is the largest multi-view dataset for clinical applications, presenting 3 possible modalities, RGB-DS (color, depth and segmentation maps). Only POV-Surgery \cite{wang2023pov} contains a higher number, but with less samples per tool (29k) and just firs-person view.


\begin{table}[]
\resizebox{1\textwidth}{!}{
\begin{tabular}{c|cccccc}
\hline
Dataset                                                       & \# frames & \begin{tabular}[c]{@{}c@{}}Source \\ (Real/ Synth)\end{tabular} & \begin{tabular}[c]{@{}c@{}}Viewpoints\\ (Single/Multi/Ego)\end{tabular} & \multicolumn{1}{c|}{Annotations} & Modalities & \begin{tabular}[c]{@{}c@{}}Clinical\\ (no. of tools)\end{tabular} \\ \hline
HO-3D \cite{Hampali2020HOnnotate}            & 77.5k     & Real                                                            & Single                                                                  & automatic                        & RGB        & -                                                                 \\
ObMan \cite{hasson19_obman}                 & 153k      & Synth                                                           & Multi                                                                   & automatic                        & RGB-DS     & -                                                                 \\
ContactPose \cite{Brahmbhatt2020ContactPose} & 2.9M      & Real                                                            & Multi                                                                   & semi-automatic                   & RGB-D      & -                                                                 \\
Hein et al. \cite{Hein2021}                  & 10.5k     & Synth                                                           & Ego                                                                     & automatic                        & RGB-DS     & 1                                                                 \\
POV-Surgery \cite{wang2023pov}                & 88k       & Synth                                                           & Ego                                                                     & automatic                        & RGB-DS     & 3                                                                 \\
\textbf{HUP-3D (ours)}                                        & 31680     & Synth                                                           & Multi                                                                   & automatic                        & RGB-DS     & 1                                                                 \\ \hline
\end{tabular}
}
\caption{Dataset comparison: HUP-3D outstands as the first multi-view 3D hand-(clinical)object dataset.}
\label{tab:hand_grasp_datasets}
\end{table}

\section{Experiment}
To support the utility of our proposed dataset HUP-3D, we deploy a deep learning (DL) state-of-the-art model designed for other datasets like HO-3D \cite{Hampali2020HOnnotate}. As mentioned before, our dataset consists of 31,680 image sets from 11 realistic hand-object grasps. In a supervised learning setting, we further split the data as 7 grasps for training (20,160), 2 grasps for validation and 2 more for testing (5,760). This will ensure the generalisation capability of the tested DL model.  
\subsection{3D hand-probe pose estimation}
There has been extensive DL methods proposed for 3D hand-object pose estimation in the computer vision community. One of these competitive baselines is HOPE-net \cite{hope}, originally tested on real data. HOPE-net extends the capabilities of residual convolutional neural networks \cite{resnet} with an adaptive Graph U-Net module \cite{graphunet}. This module manages to reduce the highly non-linear regression of the 3D hand and object coordinates.

Therefore, the task of the estimator is to map the RGB images to 3D world coordinates of the hand skeleton and the object's boundary corners. For this, we minimise a mean square error loss during training on both 2D and 3D coordinates. In terms of training, we follow the same settings as in the original HOPE-net paper \cite{hope}.

\begin{figure}[htbp]
\centering
\includegraphics[width=.9\linewidth]{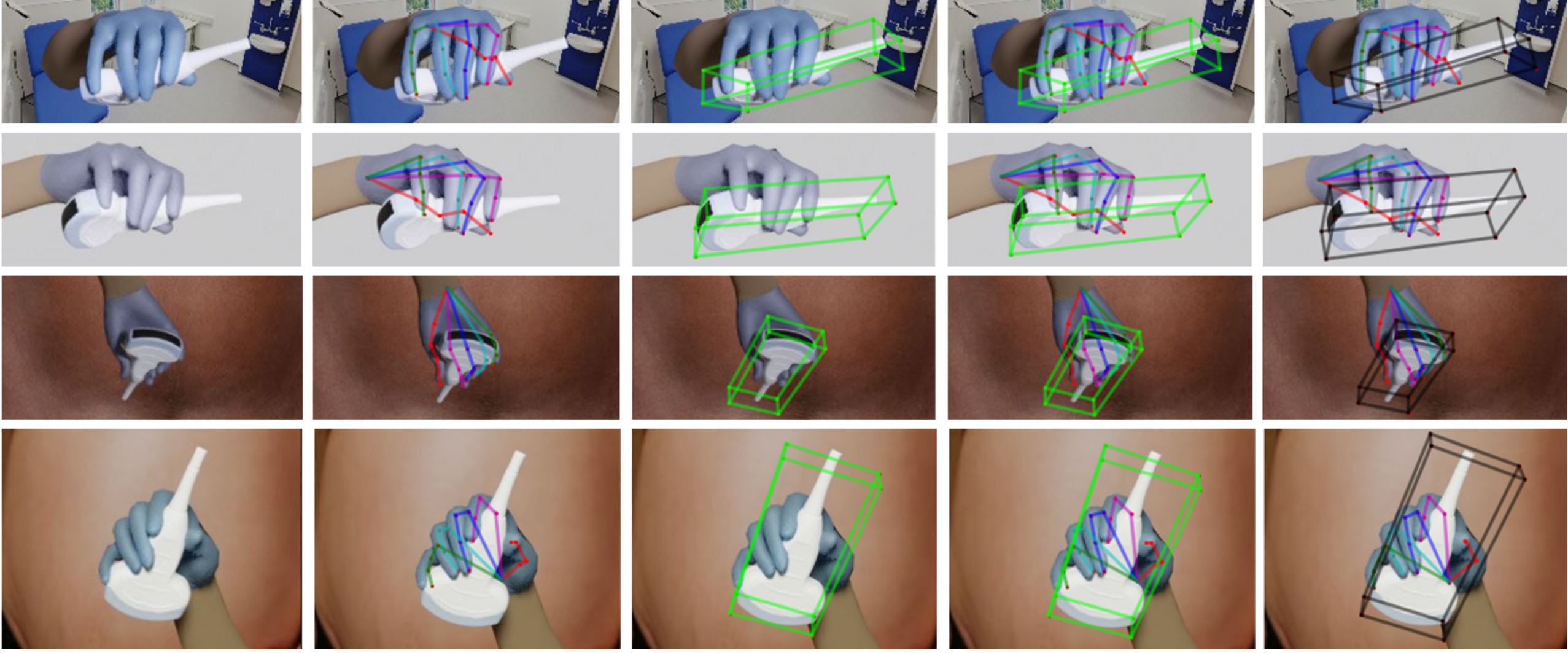}
\caption{Qualitative results, shown with 4 test images from HUP-3D: image columns from left to right: RGB, predicted hand joints, predicted probe corners, predicted joints and corners, ground truth of joints and corners}
\label{fig:qual_results}
\end{figure}

Quantitatively, once trained, we measure the error in millimeters between the predicted joints and the ground-truth. On the testing set, we obtain a total error of \textbf{8.65 mm} as a mean per joint position error (MPJPE) with 5.33 mm from the hand and 17.05 mm from the object. The testing error is the lowest compared to the other clinical datasets like POV-Surgery \cite{wang2023pov} (14.35 mm) and Hein et al. \cite{Hein2021} (17.02 mm) where even more advanced DL models were used. This proves that a multi-viewpoint dataset helps to estimate the object and the hand location with higher precision. In our experiments, we have also tested a more simple baseline composed just of ResNet-50 \cite{resnet}, where the error was higher at 9.69 mm.
In Fig. \ref{fig:qual_results}, we visually confirm the accurate predicted keypoints from our HUP-3D test set.

\section{Conclusion and future work}
We introduced HUP-3D, a pioneering multi-view 3D hand-object dataset tailored for obstetrics ultrasound hand-probe grasps. HUP-3D aims to enhance research in clinical movement analysis via egocentric camera and mixed reality applications. Our data generation process leverages a versatile model for grasp generation and an efficient automated rendering pipeline, illustrating the benefits of our multi-view camera sphere approach. A baseline model evaluation confirmed our method's effectiveness, even with significant hand-probe occlusions. Future efforts will focus on improving real-world applicability by incorporating automatically annotated real images and developing more sophisticated grasp generation techniques that incorporate temporal sequences for better manual interaction and predictions.

\clearpage
\setcounter{page}{1}
\setcounter{section}{0}
\setcounter{table}{0}
\section{Supplementary material}
\subsubsection{HUP-3D Dataset}
Our HUP-3D dataset can be downloaded via the following google drive link: \href{https://drive.google.com/file/d/1_MDn7AaansvGdU_wd_eiFO4n95R-Ri9L/view?usp=sharing}{HUP-3D}

\subsubsection{Grasp generation and rendering pipeline configuration}
Table \ref{tab:frame_generation_config} outlines the different parameters our frame generation pipeline utilizes, which influence the diversity and quantity of frames produced:
\begin{table}
\caption{Frame generation configuration}\label{tab:frame_generation_config}
\begin{tabular}{|c|m{4.8cm}|m{5.5cm}|}
\hline
\textbf{Parameter} & \textbf{Description} & \textbf{Values} \\ \hline
 $(\Theta_{Vol})_{j}$ & k-th ($k \in [1, .., N]$) probe rotation Euler angles, used within GrabNet & Manually selected set of three Euler angles $[\alpha, \beta, \gamma | \alpha, \beta, \gamma \in [0, ..., 360]]$  \\ \hline
$(G_{[\Theta_{Vol}]})_k$ & k-th manually selected GrabNet-based grasp out of all grasps generated by $\Theta_{Vol}$ & 11 manually selected plausible grasps via visual inspection in MeshLab\\ \hline
$r_{sph}$ & Sphere radius &  0.8m \\ \hline
$r_{circ}$ & Sphere surface circle radius & 0.15m \\ \hline
$\Theta_k$ & k-th ($k \in [1, .., 92]$) camera view point Euler angles that depend on $r_{sph}$ and $r_{circ}$ & 92 - 2 excluded = 90 values. Concrete Euler angles can be derived from applying the sphere concept described in section \ref{sec:sphere_concept} \\ \hline
$z_{ego}$ & Camera distance (egocentric \newline head to hand distance) & $[0.5, 0.8]$ \\ \hline
$\mathcal{G}_{bkgr}$ & Background image for rendered \newline grasp frames & 1 x plain white, 1 x consultation room, 3 x SPACE-FAN phantom, 3 x real pregnant mother belly (white / brown / black)\\ \hline
$H_{rgb}$ & RGB values of glove texture & 
$\left[\begin{array}{c}
0.5647058824 \\
0.5921568627 \\
0.768627451
\end{array}\right]$,
$\left[\begin{array}{c}
0.38039215686 \\
0.61960784314 \\
0.8666666667
\end{array}\right]$
\\ \hline
$A_{rgb}$ & RGB values of the arm &
$[1.0, 0.6784313725, 0.3764705882]$ \\
\hline
\end{tabular}
\end{table}

Using the variable notations of Table~\ref{tab:frame_generation_config}, the frame generation in our dataset can be expressed as in Algorithm~\ref{alg:combinatorics_frame_generation} below:

\begin{algorithm}[H]
\scriptsize
\caption{Combinatorics for frame generation based on a selected grasp}
\label{alg:combinatorics_frame_generation}
\begin{algorithmic}[1]
\Statex \textbf{Input:} Selected grasp $(G_{[\Theta_{Vol}]})_k$, egocentric viewpoint $z_{ego}$, background image $\mathcal{G}_{bkgr}$, hand texture $H_{rgb}$, camera view Euler angles $\Theta_k$, sphere radius $r_{sph}$, circle radius $r_{circ}$
\Statex \textbf{Output:} Set of frames $F$
\State Initialize $F \gets \emptyset$
\ForAll{$(G_{[\Theta_{Vol}]})_k, z_{ego}, \mathcal{G}_{bkgr}, H_{rgb}, \Theta_k$}
    \State $frame \gets GenerateFrame((G_{[\Theta_{Vol}]})_k, z_{ego}, \mathcal{G}_{bkgr}, H_{rgb}, \Theta_k, r_{sph}, r_{circ})$
    \State $F \gets F \cup \{frame\}$
\EndFor
\end{algorithmic}
\end{algorithm}


\clearpage
\begin{thebibliography}{8}

\bibitem{Hasson2020Leveraging}
Yana Hasson, Bugra Tekin, Federica Bogo, Ivan Laptev, Marc Pollefeys, and Cordelia Schmid.
\newblock Leveraging Photometric Consistency Over Time for Sparsely Supervised Hand-Object Reconstruction.
\newblock In \textit{Proceedings of the IEEE/CVF CVPR}, June 2020.


\bibitem{jiang2021graspTTA}
H. Jiang, S. Liu, J. Wang, and X. Wang, ``Hand-Object Contact Consistency Reasoning for Human Grasps Generation,'' in \textit{Proceedings of the ICCV}, 2021.

\bibitem{Akin2012Enhanced}
Abdulkadir Akin, E. Erdede, Hossein Afshari, Alexandre Schmid, and Yusuf Leblebici.
\newblock Enhanced Omnidirectional Image Reconstruction Algorithm and Its Real-Time Hardware.
\newblock In \textit{Proceedings - 15th Euromicro Conference on Digital System Design, DSD 2012}, Sep 2012.
\newblock \doi{10.1109/DSD.2012.52}.
\newblock ISBN 978-1-4673-2498-4.

\bibitem{Liu2021SemiSupervised}
Shaowei Liu, Hanwen Jiang, Jiarui Xu, Sifei Liu, and Xiaolong Wang.
\newblock Semi-Supervised 3D Hand-Object Poses Estimation With Interactions in Time.
\newblock In \textit{Proceedings of the IEEE/CVF CVPR}, pages 14687--14697, June 2021.

\bibitem{Oberweger2020Generalized}
Markus Oberweger, Paul Wohlhart, and Vincent Lepetit.
\newblock Generalized Feedback Loop for Joint Hand-Object Pose Estimation.
\newblock \textit{IEEE Transactions on Pattern Analysis and Machine Intelligence}, 42(8):1898--1912, 2020.
\newblock \doi{10.1109/TPAMI.2019.2907951}.

\bibitem{hasson19_obman}
Y. Hasson, G. Varol, D. Tzionas, I. Kalevatykh, M. J. Black, I. Laptev, and C. Schmid, "Learning joint reconstruction of hands and manipulated objects," in \textit{Proceedings of the IEEE Conference on Computer Vision and Pattern Recognition (CVPR)}, 2019.

\bibitem{Hein2021}
Hein, J., Seibold, M., Bogo, F., Farshad, M., Pollefeys, M., Fürnstahl, P. and Navab, N. (2021). Towards markerless surgical tool and hand pose estimation. International Journal of Computer Assisted Radiology and Surgery, 16(5), 799–808. https://doi.org/10.1007/s11548-021-02369-2

\bibitem{wang2023pov}
R. Wang, S. Ktistakis, S. Zhang, M. Meboldt, and Q. Lohmeyer, "POV-Surgery: A Dataset for Egocentric Hand and Tool Pose Estimation During Surgical Activities," in \textit{Proceedings of the International Conference on Medical Image Computing and Computer-Assisted Intervention}, 2023, pp. 440--450.

\bibitem{GRAB:2020}
O. Taheri, N. Ghorbani, M. J. Black, and D. Tzionas, ``{GRAB}: A Dataset of Whole-Body Human Grasping of Objects,'' in \textit{Proceedings of the European Conference on Computer Vision (ECCV)}, 2020. [Online]. Available: \url{https://grab.is.tue.mpg.de}

\bibitem{Hampali2020HOnnotate}
Shreyas Hampali, Mahdi Rad, Markus Oberweger, and Vincent Lepetit.
\newblock HOnnotate: A Method for 3D Annotation of Hand and Object Poses.
\newblock In \textit{Proceedings of the IEEE/CVF Conference on Computer Vision and Pattern Recognition (CVPR)}, June 2020.

\bibitem{Brahmbhatt2020ContactPose}
Samarth Brahmbhatt, Chengcheng Tang, Christopher D. Twigg, Charles C. Kemp, and James Hays.
\newblock ContactPose: A Dataset of Grasps with Object Contact and Hand Pose.
\newblock In Andrea Vedaldi, Horst Bischof, Thomas Brox, and Jan-Michael Frahm, editors, \textit{Computer Vision -- ECCV 2020}, pages 361--378. Springer International Publishing, Cham, 2020.
\newblock ISBN 978-3-030-58601-0.

\bibitem{Doosti2020HOPENet}
Bardia Doosti, Shujon Naha, Majid Mirbagheri, and David J. Crandall.
\newblock HOPE-Net: A Graph-Based Model for Hand-Object Pose Estimation.
\newblock In \textit{Proceedings of the IEEE/CVF Conference on Computer Vision and Pattern Recognition (CVPR)}, June 2020.

\bibitem{kwon2021h2o}
T. Kwon, B. Tekin, J. Stühmer, F. Bogo, and M. Pollefeys, ``H2O: Two hands manipulating objects for first person interaction recognition,'' in \emph{Proceedings of the IEEE/CVF International Conference on Computer Vision}, pp. 10138--10148, 2021.




\bibitem{miller2004graspit}
Andrew T. Miller and Peter K. Allen.
\newblock Graspit! A Versatile Simulator for Robotic Grasping.
\newblock \textit{IEEE Robotics \& Automation Magazine}, 11(4):110--122, 2004.
\newblock IEEE.

\bibitem{Blender2018}
Blender Online Community.
\newblock \textit{Blender - a 3D modelling and rendering package}.
\newblock Stichting Blender Foundation, Amsterdam, 2018.

\bibitem{Romero2017EmbodiedHands}
Javier Romero, Dimitrios Tzionas, and Michael J. Black.
\newblock Embodied hands: modeling and capturing hands and bodies together.
\newblock \textit{ACM Transactions on Graphics}, 36(6):245, November 2017.
\newblock Association for Computing Machinery, New York, NY, USA.
\newblock ISSN 0730-0301.

\bibitem{Dromey2020DimensionlessUltrasound}
B. P. Dromey, S. Ahmed, F. Vasconcelos, E. Mazomenos, Y. Kunpalin, S. Ourselin, J. Deprest, A. L. David, D. Stoyanov, and D. M. Peebles, ``Dimensionless squared jerk: An objective differential to assess experienced and novice probe movement in obstetric ultrasound,'' \emph{Prenatal Diagnosis}, vol. 11, 2020.

\bibitem{CAI2020101762}
Y. Cai, R. Droste, H. Sharma, P. Chatelain, L. Drukker, A. T. Papageorghiou, and J. A. Noble, ``Spatio-temporal visual attention modelling of standard biometry plane-finding navigation,'' \emph{Medical Image Analysis}, vol. 65, 2020.

\bibitem{prokudin2019efficient}
Sergey Prokudin, Christoph Lassner, and Javier Romero.
\newblock Efficient learning on point clouds with basis point sets.
\newblock In \emph{Proceedings of the IEEE/CVF ICCV}, 2019.

\bibitem{varol2017learning}
G. Varol, J. Romero, X. Martin, N. Mahmood, M.J. Black, I. Laptev, and C. Schmid.
\newblock Learning from synthetic humans.
\newblock In: \emph{Proceedings of the IEEE CVPR}, 2017.


\bibitem{azari2019using}
D. P. Azari, Y. H. Hu, B. L. Miller, B. V. Le, and R. G. Radwin, ``Using surgeon hand motions to predict surgical maneuvers,'' \emph{Human Factors}, vol. 61, 2019, SAGE Publications Sage CA: Los Angeles, CA.


\bibitem{8610321}
X.-H. Zhou, G.-B. Bian, X.-L. Xie, Z.-G. Hou, X. Qu, and S. Guan, ``Analysis of Interventionalists’ Natural Behaviors for Recognizing Motion Patterns of Endovascular Tools During Percutaneous Coronary Interventions,'' \emph{IEEE Transactions on Biomedical Circuits and Systems}, vol. 13, 2019.

\bibitem{droste2020automatic}
R. Droste, L. Drukker, A. T. Papageorghiou, and J. A. Noble, ``Automatic probe movement guidance for freehand obstetric ultrasound,'' in \emph{MICCAI 2020: 23rd International Conference}, Springer, 2020.

\bibitem{goodman2021real}
E. D. Goodman, K. K. Patel, Y. Zhang, W. Locke, C. J. Kennedy, R. Mehrotra, S. Ren, M. Y. Guan, M. Downing, H. W. Chen, et al., ``A real-time spatiotemporal AI model analyzes skill in open surgical videos,'' \emph{arXiv preprint arXiv:2112.07219}, 2021.


\bibitem{Jin2018ToolDetection}
A. Jin et al., ``Tool Detection and Operative Skill Assessment in Surgical Videos Using Region-Based Convolutional Neural Networks,'' in \emph{2018 IEEE WACV}, 2018.

\bibitem{lajko2021endoscopic}
G. Lajkó, R. Nagyné Elek, and T. Haidegger, ``Endoscopic image-based skill assessment in robot-assisted minimally invasive surgery,'' \emph{Sensors}, vol. 21, MDPI.

\bibitem{nguyen2022holous}
T. Nguyen, W. Plishker, A. Matisoff, K. Sharma, and R. Shekhar, ``HoloUS: Augmented reality visualization of live ultrasound images using HoloLens for ultrasound-guided procedures,'' \emph{International Journal of Computer Assisted Radiology and Surgery}, Springer vol. 17, 2022.

\bibitem{MANO:SIGGRAPHASIA:2017}
J. Romero, D. Tzionas, and M. J. Black, ``Embodied Hands: Modeling and Capturing Hands and Bodies Together,'' \emph{ACM Transactions on Graphics, (Proc. SIGGRAPH Asia)}, Nov. 2017.

\bibitem{resnet}
Kaiming He and Xiangyu Zhang and Shaoqing Ren and Jian Sun, ``Deep Residual Learning for Image Recognition'', 2015, Tech Report, eprint=1512.03385.

\bibitem{hope}
B. Doosti, S. Naha, M. Mirbagheri and D. Crandall, ``HOPE-Net: A Graph-based Model for Hand-Object Pose Estimation'', (CVPR), June, 2020.

\bibitem{graphunet}
H. Gao and S. Ji, ``Graph U-Nets", ICML, 2019 

%
%
\end{thebibliography}
\end{document}